\documentclass[runningheads]{llncs}

\usepackage{eccv}

\usepackage{eccvabbrv}

\usepackage{graphicx}
\usepackage{booktabs}
\usepackage{lipsum}
\usepackage{amsmath}
\usepackage{tabularx, makecell, multirow} 
\usepackage{gensymb}
\usepackage{enumitem}
\usepackage{pifont}
\usepackage{wrapfig,lipsum,booktabs}
\newcolumntype{x}[1]{>{\centering\let\newline\\\arraybackslash\hspace{0pt}}p{#1}}

\usepackage[accsupp]{axessibility}  

\usepackage[colorlinks]{hyperref}

\usepackage{orcidlink}

\makeatletter
\def\thanks#1{\protected@xdef\@thanks{\@thanks
        \protect\footnotetext{#1}}}
\makeatother

\usepackage{colortbl}
\definecolor{yellow}{rgb}{1, 1, 0.7}
\definecolor{orange}{rgb}{1, 0.85, 0.7}
\definecolor{tablered}{rgb}{1, 0.7, 0.7}
\newcommand{\best}{\cellcolor{tablered}}
\newcommand{\sbest}{\cellcolor{orange}}
\newcommand{\tbest}{\cellcolor{yellow}}

\begin{document}

\title{VEGS: View Extrapolation of Urban Scenes in 3D Gaussian Splatting using Learned Priors}

\titlerunning{View Extrapolation of Urban Scenes in 3D Gaussian Splatting}

\author{Sungwon Hwang\inst{1}*\thanks{* Authors contributed equally to this work.}\orcidlink{0000-0001-6688-3967} \and
Min-Jung Kim\inst{1}*\orcidlink{0000-0003-3799-8225} \and
Taewoong Kang\inst{1}\orcidlink{0009-0001-3985-8384} \and \\
Jayeon Kang \inst{2}\orcidlink{0009-0006-5653-0571} \and
Jaegul Choo \inst{1}\orcidlink{0000-0003-1071-4835}\\
}
\authorrunning{S.~Hwang and M.~Kim et al.}

\institute{~\inst{1}KAIST \ \ ~\inst{2}Ghent University
\\
\email{\{shwang.14, emjay73, keh0t0, jchoo\}@kaist.ac.kr, \\ jayeon.kang@ghent.ac.kr}
}

\maketitle
\vspace{-0.2cm}
\begin{abstract}
    Neural rendering-based urban scene reconstruction methods commonly rely on images collected from driving vehicles with cameras facing and moving forward. Although these methods can successfully synthesize from views similar to training camera trajectory, directing the novel view outside the training camera distribution does not guarantee on-par performance. In this paper, we tackle the Extrapolated View Synthesis (EVS) problem by evaluating the reconstructions on views such as looking left, right or downwards with respect to training camera distributions. To improve rendering quality for EVS, we initialize our model by constructing dense LiDAR map, and propose to leverage prior scene knowledge such as surface normal estimator and large-scale diffusion model. Qualitative and quantitative comparisons demonstrate the effectiveness of our methods on EVS. To the best of our knowledge, we are the first to address the EVS problem in urban scene reconstruction. Link to our project page: 
    \href{https://vegs3d.github.io/}{https://vegs3d.github.io/}.

\end{abstract}


\section{Introduction}
\label{sec:intro}

Advancements in neural implicit representations and their rendering methods such as NeRF\cite{mildenhall2020nerf} have enabled accurate, high-fidelity reconstruction of 3D scene and novel view synthesis \cite{barron2021mip, barron2022mip, barron2023zip, muller2022instant}. However, these methods assume certain conditions such as staticity of scene, or dense and diversely distributed training images for accurate scene reconstruction. To handle non-static scenes, a line of works \cite{pumarola2021d, park2021nerfies, park2021hypernerf} define canonical space and temporal latent vectors to encode per-frame deformation, or learn to separate transient objects via space uncertainty modeling \cite{martin2021nerf, sun2022neuconw}. 
To relax the dense training set requirements, 
 various methods have been proposed to train NeRFs given a few sparsely distributed images \cite{yu2021pixelnerf, jain2021putting, niemeyer2022regnerf, yang2023freenerf}. However, these works mainly focus on the small number of training cameras rather than their pose distribution, which can also be problematic when it is biased toward a certain location or viewpoint.
 



\begin{center}
    \begin{figure}[t]
        \includegraphics[width=\textwidth]{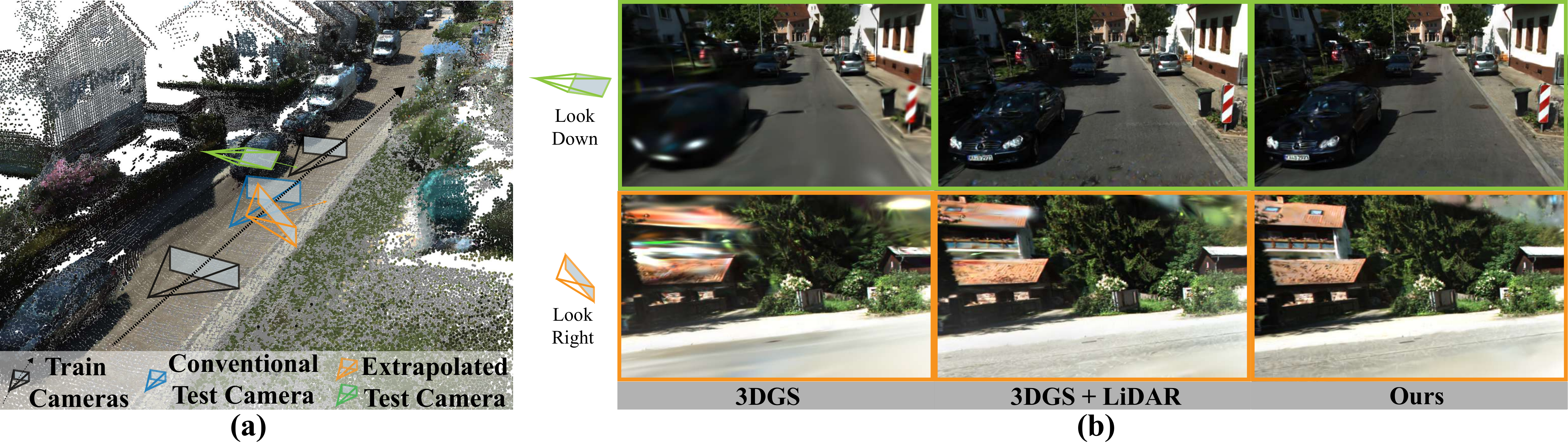}
        \captionof{figure}{\textcolor{black}{(a) Illustration of Extrapolated View Synthesis (EVS) problem in urban scenes reconstructed with forward-facing cameras.     
        In contrast to conventional test cameras similar to training camera poses, we evaluate view synthesis on cameras distant from training camera distribution. 
        (b) Qualitative comparison on EVS to baselines.}}
        \label{fig:teaser}
    \end{figure}
\end{center}

Meanwhile, some other methods raised specific solutions for urban scene reconstruction using NeRF-based methods. Most of these works either focus on reconstructing scenes with dynamic objects \cite{ost2021neural, wu2023mars, fu2022panoptic} or improving modeling capacity \cite{ost2021neural, Turki_2022_CVPR}, as urban scenes tend to be in large-scale. Notably, Neural Scene Graph \cite{ost2021neural} and MARS \cite{wu2023mars} propose to model urban scenes with a graph that comprises multiple neural implicit models for static and dynamic objects as nodes, and 3D bounding boxes and their spatial relations as edges, followed by demonstrating their methods on common driving scene dataset such as KITTI\cite{Geiger2012CVPR}. Block-NeRF \cite{tancik2022block} proposes to effectively model large-scale scene by dividing a space into multiple blocks, each of which is represented with an independent NeRF network.

However, none of the existing methods on urban scenes address the limited view distribution of training images commonly collected from cameras on vehicles facing and moving forward. Since such characteristic is quite contrary to requiring diversely posed images for accurate scene reconstruction\cite{mildenhall2020nerf}, one can easily insinuate that rendering from viewpoints far-distanced from training cameras may yield lower quality. In fact, existing works on urban scene reconstruction \cite{ost2021neural, fu2022panoptic, wu2023mars} construct training and test viewpoints from a single set of forward-facing posed images, which makes the test viewpoints to reside in "\textit{interpolative}" area defined by training cameras. Thus, evaluation on these test cameras is irrelevant for view synthesis looking far on the left, right, and downward with respect to the distribution of training cameras. Considering that observation from such extrapolated views is essential for maximal use of reconstructed scenes, we intend to focus our work on observing, analyzing, and improving rendering quality from these views.

As shown in Figure \ref{fig:teaser}, we formulate such problem as Extrapolated View Synthesis (EVS), and demonstrate that rendering quality does degrade on EVS over existing methods even when they render successfully on the interpolative test cameras. To address the problem, we propose three methods to improve rendering quality on EVS by distilling prior knowledge from LiDAR, surface normal estimator, and large-scale image diffusion model to our scene reconstructions. 

Since many applications of view synthesis on urban scene require real-time view synthesis \cite{kaur2021survey}, we stem our method from 3D Gaussian Splatting \cite{kerbl20233d}, a point-based scene representation method that can yield high-quality rendering in real time with $\approx 144$ fps. We propose a method to model and initialize a dynamic scene given point-clouds from LiDAR and off-the-shelf 3D object detectors in order to guide the model with accurate geometry to improve EVS. During scene reconstruction training with photometric loss, we also propose a method to distill surface normal estimations from training images in order to shape and orient covariances of 3D Gaussians suitable for EVS. We then propose a method to fine-tune a large-scale image diffusion model to teach the visual characteristic of the scene while keeping its generalization capability for unseen views, followed by distilling that knowledge to EVS. 

In summary, the contributions of this work are four-fold:



\begin{itemize}[label=$\circ$]
    \item First to tackle extrapolated view synthesis on urban scenes reconstructed with forward-facing cameras to the best of our knowledge.
    \vspace{0.1cm}
    \item Proposal of a dynamic urban scene modeling and reconstruction method in 3D Gaussians \cite{kerbl20233d} using LiDAR. 
    \vspace{0.1cm}
    \item Proposal of a rendering and supervision method of covariances in 3D Gaussians with surface normal priors.  
    \vspace{0.1cm}
    \item Proposal of a method to training and distilling knowledge from large-scale diffusion model to unobserved views. 
\end{itemize}

\section{Related Works}
\label{sec:related}
\subsection{Neural Scene Representation}
Recent innovations driven by NeRF \cite{mildenhall2020nerf} and its variants \cite{barron2021mip, barron2022mip, barron2023zip} have enabled accurate 3D reconstruction by supervising MLP with densely posed images via differentiable volume rendering. While another line of works \cite{muller2022instant, garbin2021fastnerf} have improved the rendering speed of NeRFs, 3DGS\cite{kerbl20233d}, a unique form of point-based rendering, brought another step of innovation in terms of high-fidelity real-time rendering via point-based scene representation followed by its differential, rasterization-based splatting techniques.

As real scenes tend to be dynamic, recent works \cite{park2021nerfies, pumarola2021d, tretschk2021nonrigid} define a continuous deformation field that maps an observation coordinate to canonical coordinate where a template NeRF is defined. Notably, HyperNeRF\cite{park2021hypernerf} introduces additional high-dimensional canonical space to expand NeRF's capacity to capture topologically-varying motions. Meanwhile, scene reconstruction methods for driving scenes model dynamic objects via bounding-box detections, with an assumption that common objects in driveways such as cars are static within its bounding-box coordinate. Specially, NSG\cite{ost2021neural} proposed dynamic scene graphs to handle multiple dynamic objects in urban scenes, followed by MARS \cite{wu2023mars} with instance-aware modeling of dynamic objects.

\subsection{Scene Reconstruction with Constrained Viewpoints}
Many recent works on few-shot NeRFs defines a problem where there are a few sparsely posed yet well-distributed images for training. Some representative works employs fully convolutional networks \cite{yu2021pixelnerf}, vision transformers \cite{niemeyer2022regnerf}, normalizing flow models\cite{jain2021putting}, or diffusion models \cite{wynn2023diffusionerf} as a prior to compensate the lack of training images. 



Works closest to our problem definition tackles extrapolated view synthesis, where biased distribution of train cameras are heavily emphasized rather than their number. RapNeRF \cite{zhang2022ray} assumes training cameras to be densely posed in a certain altitude, and test their model in different altitudes. However, the method assume view-agnostic color for pseudo-guidance of unseen rays, which is inappropriate to capture outdoor scenes that often include reflective surfaces or varying lighting conditions, whose images are highly view-dependent. Conversely, NeRFVS \cite{yang2023nerfvs} enhances the approach by incorporating holistic priors, such as pseudo depth maps and view coverage, derived from neural reconstructions. The method is demonstrated for 3D indoor scenes, offering a possible solution for rendering quality across diverse appearances. Meanwhile, we tackle a new extrapolated view synthesis set-up in outdoor driving scenes where training cameras tend to face and move forwards.  


\subsection{Scene Reconstruction with Priors}

Recent works leverage geometry prior for accurate scene reconstruction. DS-NeRF \cite{kangle2021dsnerf} harnesses free depth from SfM for neural rendering, while neural RGB-D surface reconstruction\cite{Azinovic_2022_CVPR} integrates depth from RGB-D sensors into the NeRF framework for precise 3D models. Notably, MonoSDF\cite{Yu2022MonoSDF} demonstrates that depth and normal cues significantly improve reconstruction quality and optimization time. Meanwhile, many urban scene reconstruction methods leverage LiDAR, as it is a common sensor for vehicles in driving scenes. S-NeRF \cite{xie2022s} densifies per-frame sparse LiDAR scans via a depth completion network, which is used as a pseudo-guidance for depth renderings. Another LiDAR-based NeRF \cite{chang2023neural} builds a LiDAR map for scene model. However, their proposed rendering method yields sparse images, not to mention that dynamic objects such as cars that are commonly present in urban scenes are not handled.

\section{Method}
Given a sequence of frames $k \in \{1 \cdots K\}$ of dynamic urban scene images $\mathcal{I}^k$ captured from forward-facing cameras on driving vehicles, and a sequence of point-cloud set $\mathcal{P}_{k}$ collected from LiDAR sensor, our goal is to reconstruct a driving scene that can yield photo-realistic renderings on views that are not located in training cameras' distribution. In this article, we will refer to renderings on such views as Extrapolated View Synthesis (EVS). We specify how the camera poses for EVS are parameterized in \cref{sec:exp}.

\begin{figure}[t]
    \includegraphics[width=\textwidth]{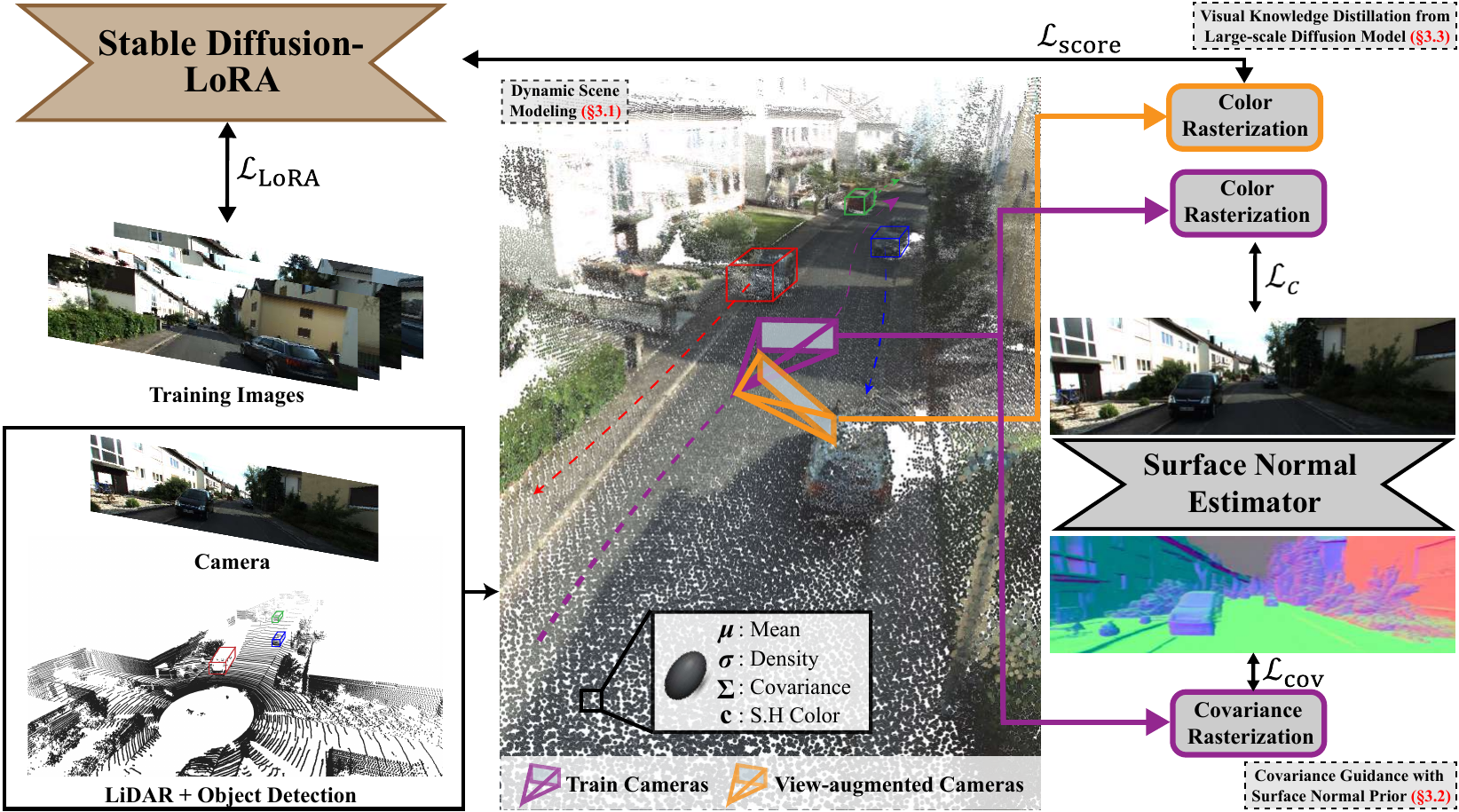}
    \caption{\textcolor{black}{Our dynamic scene model combines camera, LiDAR, and bounding box estimations with 3D Gaussian Splatting \cite{kerbl20233d} Aside from reconstruction loss $\mathcal{L}_c$, we additionally supervise Gaussian covariances with surface normal priors for improved extrapolated view synthesis (EVS). We also make use of a large-scale diffusion model to distill its knowledge directly to renderings of view-augmented cameras.}}
    \label{fig:pipeline}
\end{figure}


Our dynamic scene model integrates camera, LiDAR, and a standard bounding box estimator, leveraging 3D Gaussian Splatting \cite{kerbl20233d} to construct a static and multiple instance-wise Gaussian models (\cref{subsec:lidar}).
In addition, we learned that optimizing Gaussian models with forward-facing cameras causes the covariance shapes of Gaussians to over-fit to a certain view, making the model unsuitable for EVS. For that, we propose to guide covariance orientation and shape using surface normal priors, introducing a new covariance renderer and supervision method with surface normal maps extracted from training images (\cref{subsec:normal}).
Finally, we propose a method for directly supervising extrapolated views by distilling knowledge from a large-scale diffusion model, which we fine-tune a subset of parameters to balance between scene-specific knowledge and generalization to unseen views (\cref{subsec:diffusion}). We summarize our method in Fig. \ref{fig:pipeline}.

%



\subsection{Point-based Neural Rendering with LiDAR integration}

\label{subsec:lidar}
Previous method\cite{xie2022s} uses per-frame LiDAR scan as a sparse depth supervision. However, considering that a camera frame can also leverage scans from another frames that are visible and within view frustum, we instead propose to construct and utilize a dense point cloud map to distill concentrated scene geometry knowledge to all training views. 

\vspace{-0.3cm}
\subsubsection{Dynamic Scene Modeling and Initialization} Our dynamic scene model $M$ comprises a static model $M^{s}$ and multiple dynamic object models $M^{i}$, where $i$ refers to an $i$-th instance-wise object. Following 3D Gaussian \cite{kerbl20233d}, each model is represented with a set of Gaussian mean $\boldsymbol{\mu}$, a 3D covariance matrix $\boldsymbol{\mathrm{\Sigma}}$, density $\boldsymbol{\sigma}$, and color $\boldsymbol{c}$. Covariance matrix is further parameterized by a diagonal scaling matrix $\boldsymbol{\mathrm{S}}$ and a rotation matrix $\boldsymbol{\mathrm{R}}$, so that 

\begin{equation}
    \boldsymbol{\mathrm{\Sigma}} = \boldsymbol{\mathrm{R}}\boldsymbol{\mathrm{S}}\boldsymbol{\mathrm{S}}^{\mathrm{T}}\boldsymbol{\mathrm{R}}^{\mathrm{T}}.
    \label{eq:cov}
\end{equation}

We learned that instead of using sparse LiDAR scans as ground-truth label for optimization, initializing Gaussian means $\boldsymbol{\mu}$ with dense LiDAR maps achieves reasonable balance from over-dependence on LiDAR prior, as LiDAR scans are often prone to measurement noise~\cite{adams2000lidar}. 

Specifically, we separate per-frame LiDAR point clouds to static and instance-wise dynamic points, after which we stack each of them across frames to construct a dense static map and instance-wise point cloud objects. Formally, given $P_{k}$, we first use an off-the-shelf 3D bounding box estimator $E(\cdot)$ to yield per-instance and frame bounding box as

\begin{equation}
    b^{i}_{k} = E(P_k).
\end{equation}

\noindent Using $b^{i}_k$, we cull dynamic points within the box, and aggregate them across the frames to initialize means for each instance-wise dynamic Gaussian model, $ \boldsymbol{\mu}^{i}$, that are defined in canonical bounding-box coordinate as 

\begin{equation}
    \boldsymbol{\mu}^{i} = \oplus^{k \in K} T^{k}_{i} P^{i}_{k}, 
\end{equation}

\noindent where $P^{i}_k$ are sub-set of $P_{k}$ bounded by $b^{i}_k$, $T^{k}_{i}$ is transformation matrix from LiDAR coordinate in $k$-th frame to canonical bounding-box coordinate of $i$-th instance, and $\oplus^{k \in K}$ is concatenation across $K$ frames. We can similarly collect static scene points as
\begin{equation}
    \boldsymbol{\mu}^{s} = \oplus^{k \in K} T^{k}_{\text{w}} P^{s}_{k}, 
\end{equation}

\noindent where $P^{s}_{k}$ are sub-set of $P_{k}$ that are bounded by none of $b^{i}_k$, and $T^{k}_{\text{w}}$ is a transformation matrix from LiDAR coordinate in $k$-th frame to world coordinate. In addition, colors of all Gaussians are initialized by projecting $P_{k}$ to camera planes to assign colors. 

\vspace{-0.3cm}
\subsubsection{Dynamic Scene Rendering and Training} To render our dynamic scene, dynamic Gaussian Models in box canonical space should be mapped to world coordinate using known transformation from canonical box coordinate of $i$-th instance to bounding box location in world coordinate at $k$-th frame, $T^{i}_{k}$. That is, 

\begin{equation}
    \boldsymbol{\mu}^{i}_{k} = T^{i}_{k} \boldsymbol{\mu}^{i}, \ \ \ \textbf{R}^{i}_{k} = R^{i}_{k} \textbf{R}^{i},
\end{equation}

\noindent where $R^{i}_{k}$ is a rotation matrix of $T^{i}_{k}$, and $\textbf{R}^{i}$ is the rotation matrix that parameterizes a covariance matrix of Gaussian. Finally, all static and dynamic models in world coordinate are jointly rasterized for rendering. Specifically, Gaussian means and covariances are projected to a camera image plane to yield projected 2D mean and covariance $\mu$ and $\mathrm{\Sigma}$ using camera extrinsics $Q$, intrinsics $K$ and its jacobian $\mathrm{J}$ as

\begin{equation}
    \mu = K Q \boldsymbol{\mu}, \ \ \  \mathrm{\Sigma} = \mathrm{J} Q \boldsymbol{\mathrm{\Sigma}} Q^{\mathrm{T}} \mathrm{J}^{\mathrm{T}}.
\end{equation}

\noindent $\mu$, $\mathrm{\Sigma}$ and point density $\boldsymbol{\sigma}$ are then used to calculate the probability of rasterized Gaussian to a pixel to calculate $\alpha_{j}$ \cite{kerbl20233d}, followed by alpha blending of Gaussians for each pixel as 

\begin{align}
    \tilde{\textbf{c}} = \sum_{j \in \mathcal{N}} \textbf{c}_j \alpha_{j} \prod_{l=1}^{j-1} (1 - \alpha_{l}),
    \label{eq:c_render}
\end{align}

\noindent where $\textbf{c}_j$ is a view-dependent color calculated with spherical harmonics, and $\mathcal{N}$ are indices of ordered points that overlaps the pixel. The scene renderings are then optimized with training images using a photometric loss following \cite{kerbl20233d} as 

\begin{equation}
    \mathcal{L}_{c} = (1 - \lambda) \mathcal{L}_1 + \lambda \mathcal{L}_\text{D-SSIM}.
\end{equation}

\vspace{-0.3cm}
\subsubsection{Bounding Box Optimization} In fact, noisy bounding box estimation can cause a dynamic model to be transformed to inaccurate position in world coordinate that does not correspond to its images projected to training cameras. As so, we jointly optimize $T^{i}_{k}$, a transformation from canonical box coordinate of $i$-th instance to world coordinate at $k$-th frame, by employing an extra transformation with learnable matrix $\Delta T^{i}_{k}$ defined for every instance and frame, so that $T^{i}_{k}$ can be replaced with 

\begin{equation}
    {T'}^{i}_{k} = T^{i}_{k} \Delta T^{i}_{k},
\end{equation}

\noindent where $\Delta T^{i}_{k}$ can be further parameterized with a quaternion vector $\Delta q$ and a translation vector $\Delta t$ to constrain its optimization within geometrically plausible space. In addition, we regularize $\Delta T^{i}_{k}$ to identity transformation using the loss $\mathcal{L}_\text{box} = || \Delta q - q_{id.}||_{2} + ||\Delta t ||_{2}$, where $q_{id.}$ is an identity quaternion, so that ${T'}^{i}_{k}$ can reside around the initial estimation.


\subsection{Covariance Guidance with Surface Normal Prior}
\label{subsec:normal}

\begin{figure}[!t]
    \includegraphics[width=\textwidth]{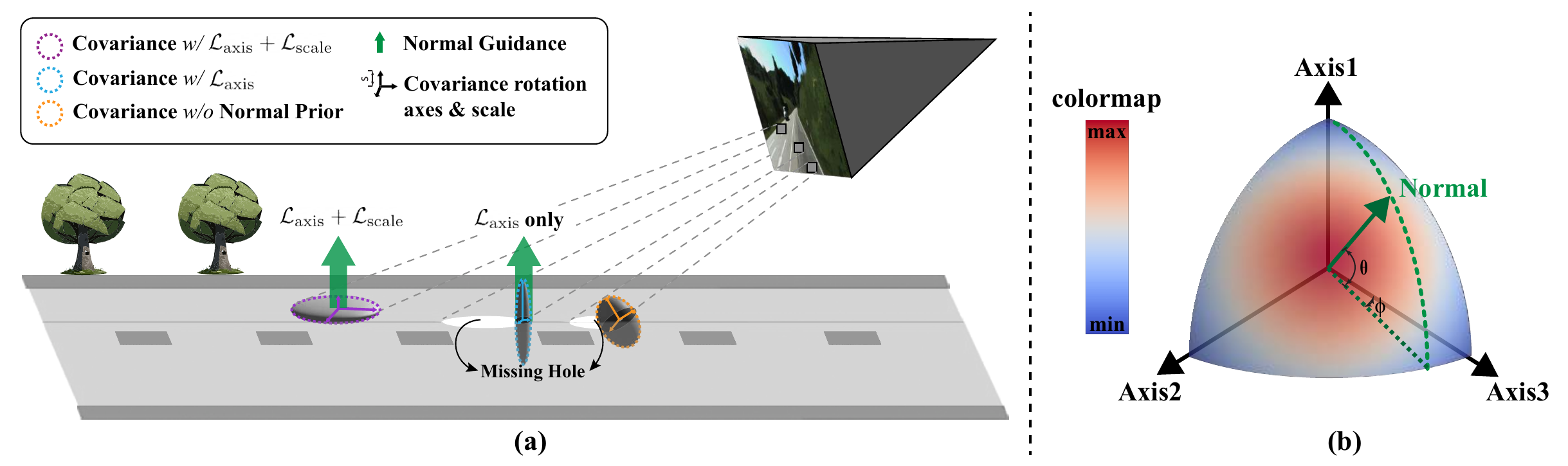}
    \caption{(a) Working mechanism of $\mathcal{L}_{\text{cov}}=\mathcal{L}_{\text{axis}} + \mathcal{L}_{\text{scale}}$. $\mathcal{L}_{\text{axis}}$ aligns covariance axes to a surface normal vector, and $\mathcal{L}_{\text{scale}}$ minimizes the scale along the covariance axis aligned with surface normal, all of which prevents the Gaussian covariance from minimally satisfying a pixel view frustum, which causes cavity when viewed from another angle. (b) Visualizing $\mathcal{L}_{\text{axis}}$ for different alignment between normal and covariances. $\mathcal{L}_{\text{axis}}$ is minimized when an axis aligns with the normal. See supplements for detailed derivation.}
    \label{fig:normal_prior}
    \vspace{-0.4cm}
\end{figure}

\subsubsection{The Lazy Covariance Optimization Problem}
In this section, we identify and tackle the limitation of a 3D Gaussian model optimized with forward-facing cameras. As illustrated in Fig. \ref{fig:normal_prior} (a), the shape and orientation of learned covariances tend to over-fit to a certain viewing angle, which we hypothesize that the covariance is trained to cover the the frustum of a training pixel with a minimal optimization effort. As a result, these covariances are prone to produce unwanted cavity on an underlying scene surface, which is revealed when viewed from unobserved angles. 

Our key idea is to guide the orientation and shape of covariances to make them behave like the underlying scene surface. In fact, unlike MLP-based representations \cite{Yu2022MonoSDF} that can calculate scene surface normal by taking negative gradient of density field with respect to a position via Autograd \cite{paszke2017automatic} library, our model cannot render a normal map due to the nature of a discrete representation of Gaussian models.  
Instead, we suggest a novel covariance rendering technique to approximate scene surface normal from rendered covariance map. Then, we guide the map with a surface normal estimated from training images in two steps: First, we align the orientation of covariances to surface normals using $\mathcal{L}_{\text{axis}}$, followed by flattening the covariance map toward the surface with $\mathcal{L}_{\text{scale}}$. The intuition behind this optimization goal is illustrated in \cref{fig:normal_prior}.

\vspace{-0.3cm}
\subsubsection{Covariance Axes Loss} 
We first propose a method to render covariances axes expressed in quaternion. As alpha-blending based on linear composition is not suitable for quaternion, we re-design~\cref{eq:c_render} to render the covariance orientation map $\tilde{\mathbf{q}}$ as
\vspace{-0.1cm}
\begin{equation}
    \tilde{\textbf{q}} = \prod_{j \in \mathcal{N}} \mathcal{S}\left( \textbf{q}_{I}, \textbf{q}_j, w_{j} \right), \ \ \ w_{j} = \alpha_{j} \prod_{l=1}^{j-1} (1 - \alpha_{l})
    \label{eq:q_render}
\end{equation}

\noindent where $\textbf{q}_{I}$ is an identity quaternion, $\mathcal{S}(\textbf{q}_{I}, \textbf{q}_{j}, w_{j})$ is a slerp function that spherically weights the orientation of $j$-th covariance $\textbf{q}_{j}$ with respect to $\textbf{q}_{I}$ by $w_{j}$. 
Weighted covariance orientations are then multiplied for cumulative application of rotations~\cite{kuipers1999quaternions}.
The rendered quaternion vector map is reformulated into a rotation matrix map and transformed into a training camera coordinate, which we denote as a covariance orientation map in matrix form $\tilde{\mathbf{Q}}$.



$\tilde{\mathbf{Q}}$ is then supervised with surface normal estimated from training images using an off-the-shelf normal prediction network $G$. Formally, our covariance axes loss is defined as, 

\begin{equation}
\mathcal{L}_{\text{axis}} = \sum\limits_{i\in \{\text{0,1,2}\}} {|\tilde{\mathbf{Q}}[:, i] \cdot G(\mathcal{I})|}/3,
\label{eq:loss_axis}
\end{equation}

\noindent where $\tilde{\mathbf{Q}}[:,i]$ represents the $i$-th axis of pixel-wise rendered covariance orientation matrix. As illustrated in \cref{fig:normal_prior} (b), $\mathcal{L}_{\text{axis}}$ is minimized when any of the three covariance axes aligns with the normal vector. We make detailed derivation of this loss in supplements. 




\vspace{-0.3cm}
\subsubsection{Covariance Scale Loss}
\label{sec:cov_scale_loss}
Axis alignment itself, however, cannot prevent the lazy covariance optimization problem, as the scale of the axis that aligns with the normal can still increase to cover the pixel view-frustum, which can still cause the cavity problem. As so, scale of the axis aligned to normal must be minimized to finally induce the covariance to mimic an underlying surface.

Specifically, we can render a covariance scale map $\tilde{\textbf{S}}$ similar to \cref{eq:c_render}, and minimize scales proportional to the cosine similarity of its axis with a normal vector. As a result, scale for normal-aligned axis will be minimized, while the remaining two scales can be trained more freely to satisfy the reconstruction loss $\mathcal{L}_{c}$. 
Formally, we establish the scale loss as

\begin{equation}
    \mathcal{L}_{\text{scale}} = \sum\limits_{i\in \{\text{0,1,2}\}} {\left|\tilde{\mathbf{S}}[i]\left(\tilde{\mathbf{Q}}[:, i] \cdot G(\mathcal{I}) \right)\right|}/3,
\label{eq:loss_scale}
\end{equation}
where $\tilde{\mathbf{S}}[i]$ is the scale of $i$-th axis of $\tilde{\mathbf{S}}$. Also, we do not back-propagate to $\tilde{\mathbf{Q}}$ in $\mathcal{L}_{\text{scale}}$ to clearly disentangle the working mechanism of $\mathcal{L}_{\text{axis}}$ and $\mathcal{L}_{\text{scale}}$. Finally, we define our covariance guidance loss as $\mathcal{L}_{\text{cov}} = \lambda_{\text{axis}} \mathcal{L}_{\text{axis}} + (1 - \lambda_{\text{axis}}) \mathcal{L}_{\text{scale}}$.

\subsection{Visual Knowledge Distillation from Large-scale Diffusion Model} 
\label{subsec:diffusion}

\subsubsection{Denoising Score Matching for Visual Knowledge Distillation} 

Apart from leveraging scene priors such as LiDAR or surface normals during optimization from training cameras, we augment cameras to EVS in order to perform direct guidance to unseen views. However, as training data is not provided for EVS, we instead make use of an image diffusion model in order to distill its knowledge on visual sanity. 

We leverage from \cite{vincent2011connection, ho2020denoising} that noise predicted from a diffusion model $\textbf{s}_{\theta}$ is proportional to the log-gradient of prior distribution, or denoising score matching given noise that is small enough \cite{song2019generative}. That is, given $\textbf{x}_{\tau} = \sqrt{\bar{\alpha}_{\tau}} \textbf{x} + (1 - \bar{\alpha_{\tau}}) \epsilon$, where $\epsilon \thicksim \mathcal{N}(0, 1)$, timestep $\tau$, pre-defined noise schedule $\bar{\alpha}_{\tau}$, and an image \textbf{x},

\begin{align}
    \textbf{s}_{\theta}(\textbf{x}_{\tau}, \tau) \approx - \nabla_{\textbf{x}} \text{log} p(\textbf{x}),
    \label{eq:score}
\end{align}

\noindent Thus, optimizing $\textbf{x}_{\tau}$ to yield smaller score pushes \textbf{x} to our prior distribution $p(\cdot)$. Similar to Perturb-and-Average Scoring in Score Jacobian Chaining (SJC) \cite{wang2023score} and DiffusioNeRF \cite{wynn2023diffusionerf}, we design our loss function using Eq.\eqref{eq:score} as 

\begin{equation}
    \nabla_{M} \mathcal{L}_{\text{score}} = - \textbf{s}_{\theta}(\hat{\mathcal{I}}_{\tau}, \tau),
\end{equation}

\noindent where $\hat{\mathcal{I}_{\tau}} = \sqrt{\bar{\alpha}_{\tau}}$ $\hat{\mathcal{I}}$ + $(1 - \bar{\alpha}_{\tau})\epsilon$ and $\hat{\mathcal{I}}$ is a rendering from our model $M$ on EVS.

\vspace{-0.3cm}
\subsubsection{Large-scale Diffusion Model with Scene-Specific Adaptation} 

Since the visual distribution of EVS is designed to resemble that of diffusion model as stated in Eq. \eqref{eq:score}, it is important for our diffusion model to have scene-specific visual understanding, yet can generalize to renderings from unseen views.

Meanwhile, recent works such as DiffusioNeRF \cite{wynn2023diffusionerf} trains DDPM\cite{ho2020denoising} with Hypersim \cite{hypersim:2021}, a synthetic indoor image dataset, in order to design a critic for visual sanity. However, guidance is conducted via 48x48 patches to prevent from over-fitting to indoor training images. As a result, the model does not strictly have scene-specific understanding, because the data used for training is not visually identical to our scene, not to mention that patch-wise supervision may not be enough to assess scene-specific visual sanity of a rendering as a whole. 
Meanwhile, GA-NeRF \cite{roessle2023ganerf} proposes GAN loss between training images and renderings from augmented views. However, adversarial training mechanism is unsuitable to our scenario due to the large difference of camera distribution between training and EVS views, making discriminator hard to be deceived. As so, adversarial training may be unsuitable for guiding unseen views. 

To satisfy both properties, we propose to fine-tune a large-scale diffusion model such as Stable Diffusion \cite{rombach2021highresolution} using LoRA \cite{hu2021lora}, a method commonly used in Large Language Models to fine-tune the low-rank residuals of projection layers in cross-attention. By doing so, our score matching model achieves generalization capability for unseen views by leveraging knowledge from large pretrained model, and scene-specific reconstruction capability by fine-tuning part of the model parameters using our training data. Formally, we use the following loss to fine-tune our diffusion model as

\begin{equation}
    \mathcal{L}_{\text{LoRA}} = \mathbb{E}_{\tau, p, \epsilon}[||\epsilon - \textbf{s}_{\theta}(\mathcal{I}_{\tau}, p)||^{2}_{2}],
    \label{eq:lora}
\end{equation}

\noindent where $p$ is a text prompt appropriately chosen for the scene, and $\mathcal{I}_{\tau} = \sqrt{\bar{\alpha}_{\tau}}$ $\mathcal{I}$ + $(1 - \bar{\alpha}_{\tau})\epsilon$ are noised training images. 

\subsubsection{Training Strategy} Prior to scene reconstruction, we first fine-tune our diffusion model $\textbf{s}_{\theta}$  using Eq. \eqref{eq:lora} using our training images. Then, we freeze $\textbf{s}_{\theta}$ and optimize our scene model $M$ using the final loss formally stated as

\begin{equation}
    \nabla \mathcal{L} = \lambda_{c} \nabla \mathcal{L}_{c} + \lambda_\text{box} \nabla \mathcal{L}_\text{box} + \lambda_{cov} \nabla \mathcal{L}_{\text{cov}} + \lambda_{\text{score}} \nabla \mathcal{L}_{\text{score}}.
\end{equation}

\begin{figure}[!t]
    \includegraphics[width=\textwidth]{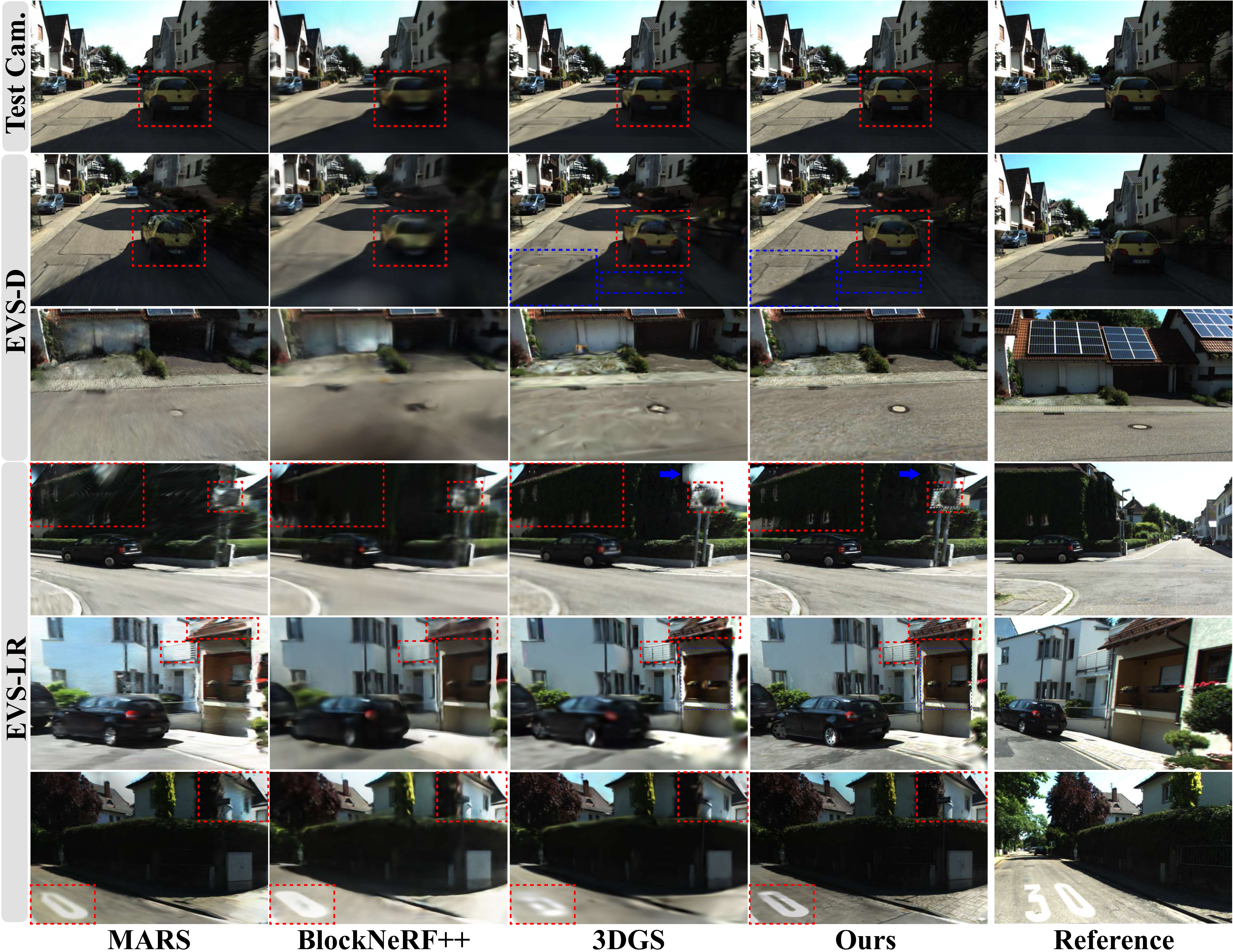}
    \caption{\textcolor{black}{Qualitative comparison on KITTI-360\cite{Liao2022PAMI} for extrapolated view synthesis. EVS-D and EVS-LR refers to extrapolated views facing downwards and left/right, respectively. Test Cam. refers to the conventional test camera sampled from a set of forward-facing cameras. We also report training images for reference that maximally covers the view space of EVS from another location for comparison. Ours outperforms the baselines in terms of geometry and visual sanity.}}
    \label{fig:quality-kitti360}
    \vspace{-0.4cm}
\end{figure}


\section{Experiments}
\label{sec:exp}

\vspace{-0.3cm}
\subsubsection{Dataset} We conduct our experiments on KITTI-360 \cite{Liao2022PAMI} and KITTI \cite{Geiger2012CVPR} Dataset. As  KITTI-360 contains 9 voluminous sequences where each sequence contains up to 15000 frames, we divide a sequence into segments of approximately 250 frames. We randomly select 16 segments with dynamic objects and another 16 segments without dynamic objects, which is for fair comparisons on EVS with baselines that do not necessarily handle dynamic objects. 

\vspace{-0.5cm}
\subsubsection{Evaluation Cameras} We first select every 8th frame as conventional test cameras. Then, we construct a EVS camera set that look left and right (EVS-LR) via rotating the test cameras by $\pm 60^{\circ}$ around Z-axis of world coordinate pointing upward, and another set that look downward (EVS-D) via rotating the test cameras by $10^{\circ}$ around the x-axis of camera coordinate pointing to the right and translating camera upward in world coordinate by $1.0$ in world unit. 
For EVS-LR, the cameras often cover under-reconstructed spaces on the side of the frame.
This nature comes from the forward-facing camera movement, which is quite common in urban scenes.
We elaborate more on this phenomenon in supplementary material. 
Since the renderings from unobserved space disturbs the quantitative results, we remove half of the image plane of EVS-LR camera farther away from the direction of train camera trajectory for experimental comparisons, and resize the cameras for EVS-D and conventional test camera to have the same image plane size with EVS-LR while keeping the same principal point.

\vspace{-0.4cm}
\subsubsection{Baselines} We made our own baseline using BlockNeRF\cite{tancik2022block}, a state-of-the-art large urban scene reconstruction method, with additional supervision with LiDAR using methods proposed by S-NeRF \cite{xie2022s} and normal loss proposed by MonoSDF \cite{Yu2022MonoSDF}, which we will denote as BlockNeRF++ in this article. We also compare our works with existing urban scene reconstruction methods such and MARS \cite{wu2023mars} that extends NSG \cite{ost2021neural} by modeling static scene with NeRF with additional depth prior supervision, as well as MipNeRF 360\cite{barron2022mip}. We also compare with 3DGS \cite{kerbl20233d} to compare relative performance between the state-of-the art point-based rendering method, and 3DGS+ that includes our dynamic scene modeling, LiDAR initialization and box optimization method to make 3DGS suitable for dynamic scenes.

\vspace{-0.3cm}
\section{Results}

\begin{figure}[!t]
    \includegraphics[width=\textwidth]{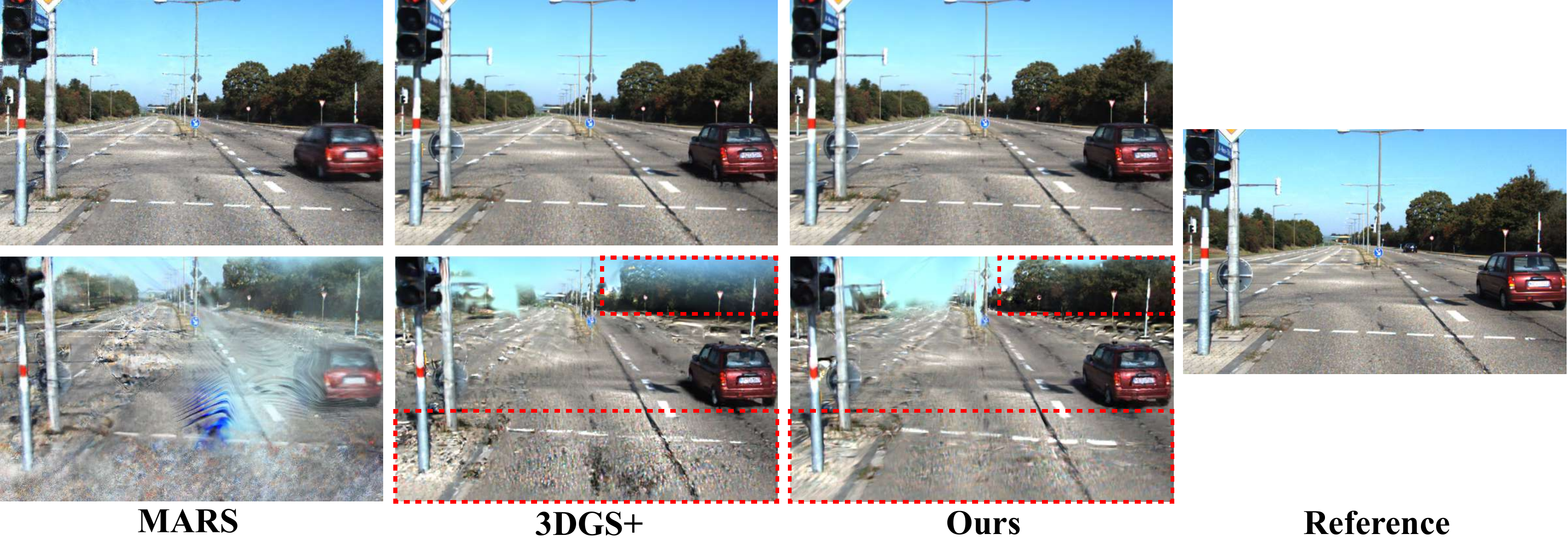}
    \caption{\textcolor{black}{Qualitative comparison on KITTI\cite{Geiger2012CVPR} dataset from conventional test camera (\textit{top}) and EVS-D (\textit{bottom}).}}
    \label{fig:quality-kitti}
\end{figure}

\begin{table}[!t]
    \centering
    \begin{tabular}{x{3.5cm}|x{1.01cm}x{1.01cm}|x{1.2cm}x{1.2cm}x{1.2cm}x{1.2cm}|x{1.01cm}}
    \toprule
         & $\text{FID} {\downarrow}$ & $\text{KID} {\downarrow}$ & $\text{PSNR} {\uparrow}$ & $\text{SSIM} {\uparrow}$ & $\text{LPIPS} {\downarrow}$ & $\text{PSNR}^{*} {\uparrow}$ & $\text{FPS} {\uparrow}$ \\
        \midrule
         Mip-NeRF 360 \cite{barron2022mip} & 181.5  & 0.1431 & 21.59 & 0.739  & 0.203 & - & 0.08  \\
         MARS \cite{wu2023mars} & \tbest 131.1  & \tbest 0.0617  & \tbest 23.13   & \best 0.814 & \sbest 0.125    &  \tbest 21.98     & 0.17   \\
         BlockNeRF++ \cite{tancik2022block, kangle2021dsnerf, Yu2022MonoSDF} &  245.1 & 0.1914 & 21.03          & 0.723          & 0.223    &   -    & 0.13   \\
         3DGS \cite{kerbl20233d} &  211.8 & 0.1382 & 21.68          & 0.772          & 0.192    &   -  & 121  \\
         3DGS+ & \sbest 126.3  & \sbest 0.0565  & \best 23.76   & \best 0.814   & \best  0.106  & \best 22.48   & 108  \\
         VEGS (\textit{ours}) &  \best 124.4 & \best 0.0561   & \sbest 23.71 & \sbest 0.812  & \best 0.106 & \sbest 22.44  & 108 \\
        \bottomrule
    \end{tabular}
    \vspace{0.2cm}
    \caption{Quantitative results on KITTI-360. FID \cite{heusel2017fid} and KID \cite{binkowski2018kid} are measured between EVS and training images. PSNR, SSIM and LPIPS are measured from conventional test cameras on static scenes where ground-truth images are available. $\text{PSNR}^{*}$ measures PSNR from conventional test cameras on dynamic object reconstructions.}
    \label{table:quantity}
    \vspace{-0.8cm}

\end{table}

\begin{figure}[!t]
    \begin{subfigure}{\linewidth}
        \centering
        \includegraphics[width=0.99\textwidth]{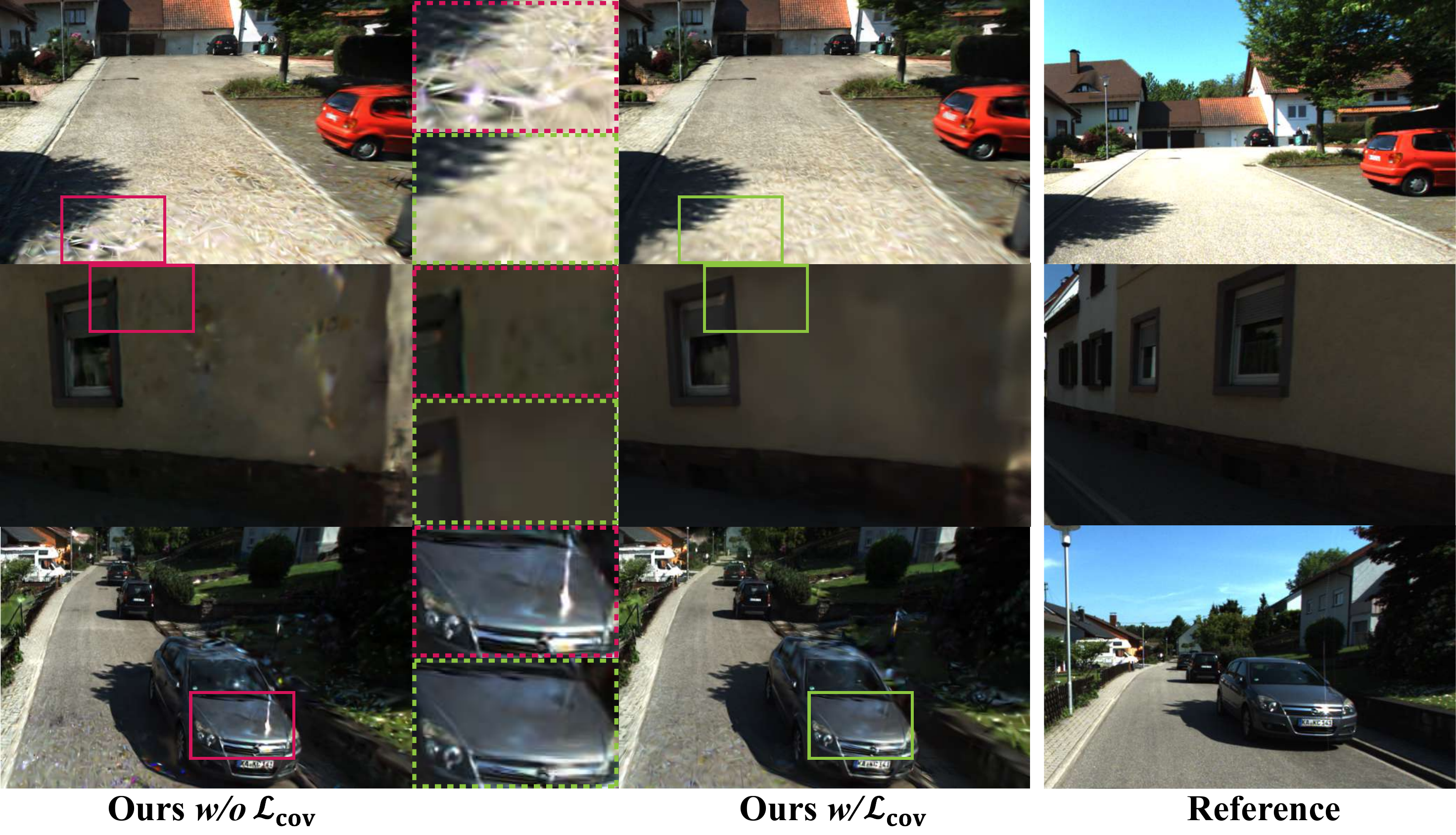}
        \caption{}
        \label{quality-kitti/ablation_normal}
    \end{subfigure}
    \begin{subfigure}{\linewidth}
        \centering
        \includegraphics[width=0.99\textwidth]{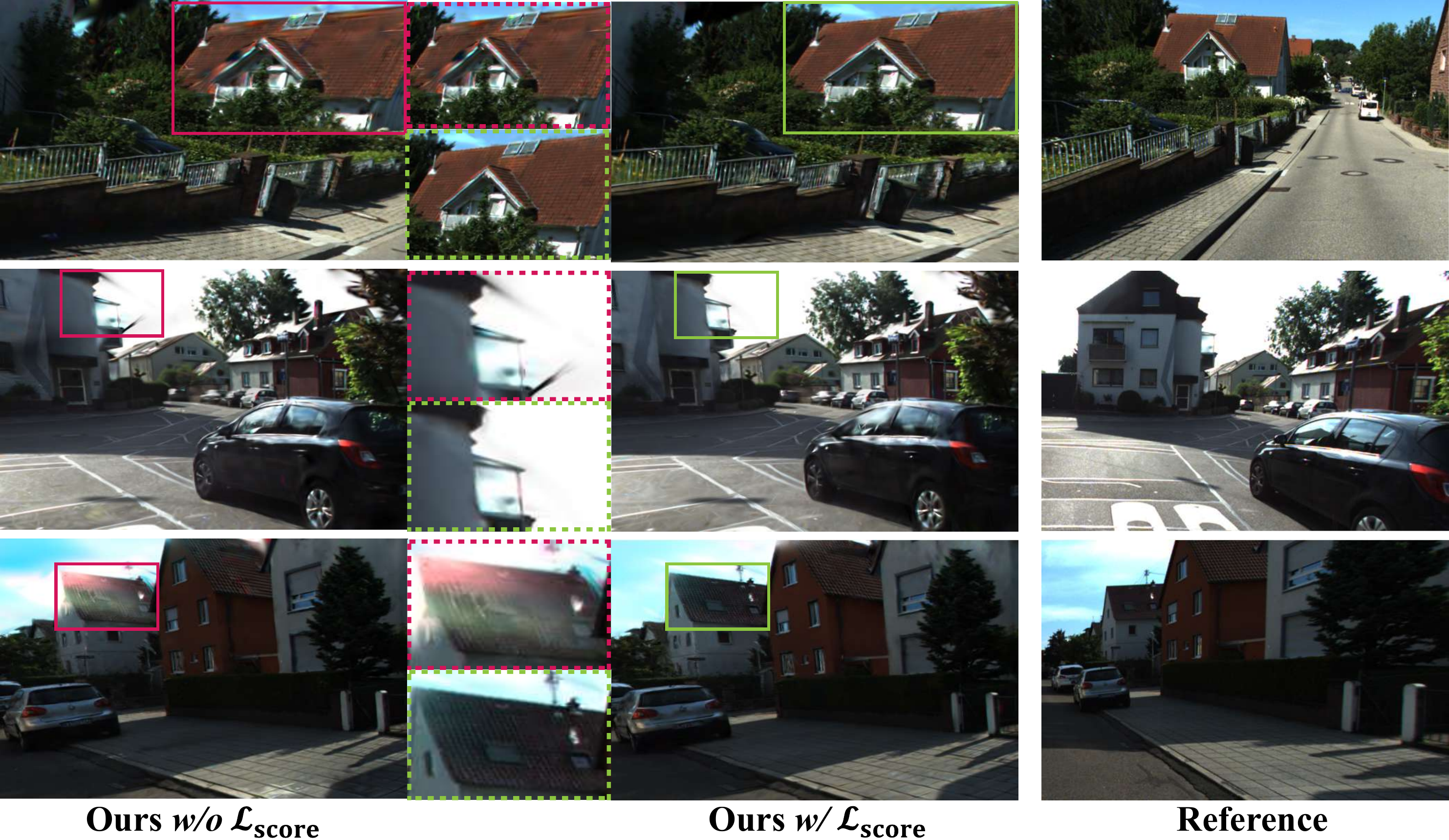}
        \caption{}  
        \label{quality-kitti/ablation_diffusion}
    \end{subfigure}
    \vspace{-0.7cm}
    \caption{Qualitative ablation results on (a) $\mathcal{L}_{\text{cov}}$ and (b) $\mathcal{L}_{\text{score}}$ on EVS. $\mathcal{L}_{\text{cov}}$ effectively guides the Gaussian covariances to faithfully cover the scene surface, yielding noticeably less cavity and better geometry. $\mathcal{L}_{\text{score}}$ effectively improves broken textures, geometry, and removes floating artifacts.}
    \vspace{-0.7cm}
\end{figure}
\vspace{-0.3cm}
\subsubsection{Comparison to Baselines} We report qualitative results of our method and baselines on KITTI-360 in \cref{fig:quality-kitti360}. Our method outperforms the baselines in both EVS-LR and EVS-D. Note that we additionally report renderings on conventional test cameras, which shows that our method is on-par with MARS and better than 3DGS and BlockNeRF++. However, comparison with MARS indicates \emph{that reconstruction quality on the conventional test cameras does not necessarily correspond to the quality on EVS}. Similar analysis can be done on qualitative results of KITTI in \cref{fig:quality-kitti}. Here, we built and compared with 3DGS+, where we included our dynamic scene modeling method with LiDAR and bounding-box detector, since SfM cannot initialize dynamic object points. 


We report quantitative results of our method with baselines in \cref{table:quantity}. FID \cite{heusel2017fid} and KID \cite{binkowski2018kid} are measured with respect to training images to measure the reconstruction qualities on EVS renderings. Even though small FID/KID cannot be expected due to the large difference of camera distribution between training images and EVS renderings, we use them as an approximation for visual sanity and closeness to the scene. We also measure PSNR, SSIM and LPIPS \cite{zhang2018perceptual} to evaluate renderings on the conventional test cameras. Ours outperforms BlockNeRF++ and 3DGS in all metrices. However, ours outperform MARS in PSNR and LPIPS, while MARS performs slightly better in SSIM, indicating that performance on conventional test cameras are on par. However, ours out-performs MARS on FID and KID measured from EVS-D and EVS-LR, which aligns with the analysis from the qualitative results in \cref{fig:quality-kitti360}. We also measure PSNR for dynamic objects only, which we denote as PSNR* in \cref{table:quantity}, and compare it with MARS. Ours yield slightly better performance in dynamic object reconstruction.

\vspace{-0.2cm}
\subsubsection{Ablations} We report qualitative ablation results on $\mathcal{L}_{\text{cov}}$ and $\mathcal{L}_{\text{score}}$ in \cref{quality-kitti/ablation_normal} and \cref{quality-kitti/ablation_diffusion}, respectively. As can be seen, the lazy covariance optimization problem is effectively ameliorated with $\mathcal{L}_{\text{cov}}$ by removing cavities on surfaces such as floor, wall, and car hood. In addition, $\mathcal{L}_{\text{score}}$ brings noticeable improvement in visual quality such as refining broken texture, geometry, and floater that we conjecture to be originated from Gaussians of ill-posed space such as sky. We report quantitative ablation results in supplements.




\vspace{-0.2cm}
\subsubsection{Scene Editing}
\begin{figure}[!t]
    \centering
    \includegraphics[width=\textwidth]{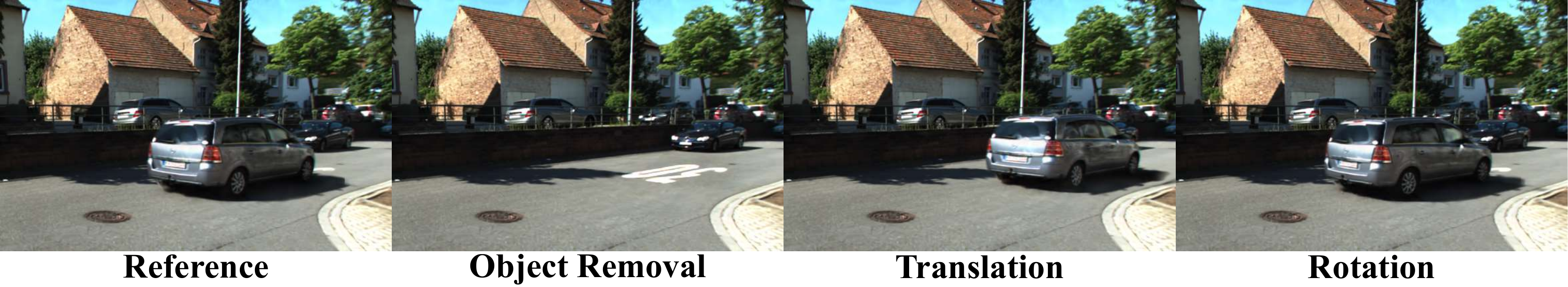}
    \vspace{-0.6cm}
    \caption{\textcolor{black}{Scene editing results. Since our method models dynamic objects on its own canonical space separated from world coordinate, the reconstructed object can be relocated or removed by manual adjustments.}}
    \label{fig:edit}

\end{figure}

In order to demonstrate the effectiveness of our dynamic scene modeling, we conducted scene editing experiments such as removing, translating or rotating the reconstructed dynamic object. We report our editing results in \cref{fig:edit}. The result indicates that the dynamic object is well-modeled and separated from the static background model.

%
%

\section{Conclusion} 

This work introduces VEGS, a urban scene reconstruction method for improved Extrapolated View Synthesis (EVS) given training images from forward-facing cameras. We introduced techniques to modeling a dynamic scene in 3D Gaussians and integrating dense LiDAR map to the model. We also proposed methods to render and supervise covariances of the Gaussians with surface normal estimations to orient and shape Gaussian covariances suitable for EVS, followed by distilling knowledge from a fine-tuned image diffusion models for better visual sanity. Our comparative studies demonstrated the efficacy of our approaches in addressing the EVS problem.

\clearpage
\noindent\textbf{Acknowledgments}
This work was supported by Institute for Information \& communications Technology Promotion(IITP) grant funded by the Korea government(MSIT) (No.RS-2019-II190075 Artificial Intelligence Graduate School Program(KAIST)), the National Research Foundation of Korea (NRF) grant funded by the Korea government (MSIT) (No. NRF-2022R1A2B5B02001913), and the Air Force Office of Scientific Research under award number FA9550-23-S-0001.

\bibliographystyle{splncs04}
\bibliography{egbib}

\clearpage

 










\newcolumntype{x}[1]{>{\centering\let\newline\\\arraybackslash\hspace{0pt}}p{#1}}




%




\hypersetup{
    colorlinks,
    linkcolor={red},
    filecolor={red},
    urlcolor={red},
    citecolor={red}
}

\def\thesection{\Alph{section}}
\def\thesubsection{\thesection.\arabic{subsection}}


\begin{center}
    \Large
    \textbf{Supplementary Material}                    
\end{center}

\noindent This supplementary material provides additional results and detailed descriptions of the experimental methodologies.
\section{Ablation Study}
\subsection{Ablation Study of Normal and Diffusion Priors}
\label{sec:abl_priors}

\begin{wraptable}{r}{0.45\textwidth}
    \vspace{-0.8cm}
    \begin{tabular}{cc|cc} 
    \toprule    
    \begin{tabular}[c]{@{}c@{}}\textbf{Normal}\\\textbf{Prior}\end{tabular} & 
    \begin{tabular}[c]{@{}c@{}}\textbf{Diffusion}\\\textbf{Prior}\end{tabular} &     
    KID$\downarrow$   & FID$\downarrow$    \\ 
    \hline
    \ding{55} & \ding{55} & 0.0565 & 126.3\\
    \ding{51} & \ding{55} & 0.0564 & 124.6\\
    \ding{51} & \ding{51} & \textbf{0.0561} & \textbf{124.4}\\    
    \bottomrule
    \end{tabular}
\caption{Ablation study of our proposed method. Metrics are evaluated on the extrapolated views from the KITTI-360 dataset. Best results are highlighted in bold.}
\label{tab:ablation}
\end{wraptable}
~\cref{tab:ablation} presents an ablation study of our proposed method applied to the extrapolated view synthesis on the KITTI-360 test cameras. 
The baseline utilizes LiDAR points as the initial mean values for covariance estimation, excluding the use of surface normal and diffusion priors.
The integration of surface normal and diffusion priors has consistently improved overall performance,
as evidenced by improvements in both KID and FID metrics.
The metrics are computed as averages across the entire dataset sequence.
\subsection{Ablation Study of Normal Prior Composing Losses}
\label{sec:abl_nloss}
Here, we demonstrate an ablation study of
losses composing covariance guidance loss in~\cref{subsec:normal}. 
~\cref{tab:ablation} shows the ablation study results on KITTI-360. 
We ablated on EVS-D as Lazy Covariance Optimization (\textbf{LCO}) is more clearly observed from grounds. 
Metrics evaluated on the EVS-D from the KITTI-360 dataset demonstrate clear improvements when all the losses are utilized.

\vspace{-0.3cm}
\setcounter{table}{2}
\begin{table}[h]
  \centering
   \resizebox{\linewidth}{!}{\begin{tabular}{@{}c||c|c|c|c@{}}
    \toprule
    &
    \textcolor{red}{\ding{55}} $\mathcal{L}_{axis}$ \textcolor{red}{\ding{55}} $\mathcal{L}_{scale}$ & 
    \textcolor{teal}{\ding{51}} $\mathcal{L}_{axis}$ \textcolor{red}{\ding{55}} $\mathcal{L}_{scale}$ & 
    \textcolor{red}{\ding{55}} $\mathcal{L}_{axis}$ \textcolor{teal}{\ding{51}} $\mathcal{L}_{scale}$ & 
    \textcolor{teal}{\ding{51}} $\mathcal{L}_{axis}$ \textcolor{teal}{\ding{51}} $\mathcal{L}_{scale}$ \\
    \midrule
    FID $\downarrow$ / KID $\downarrow$ &
    $123.28 \ \ / \ \ 0.05542$  &
    $122.56 \ \ / \ \ 0.05537$  &
    $122.80 \ \ / \ \ 0.05527$  & 
    $\textbf{121.60} \ \ / \ \ \textbf{0.05521}$ \\
    \bottomrule
  \end{tabular}}
  \caption{Ablation study on $\mathcal{L}_{axis}$ and $\mathcal{L}_{scale}$.}
  \label{tab:ablation}
\end{table}
\vspace{-0.5cm}

\section{Minima Analysis of Covariance Axes Loss}
In this section, we show that the proposed covariance axes loss defined in ~\cref{eq:loss_axis} is minimized when one of the covariance axes aligns with the normal axis.
We denote the polar and azimuthal angles of the normal vector in covariance axis coordinate by $\theta$ and $\phi$, respectively.
Accordingly, ~\cref{eq:loss_axis} can be reformulated as:
\begin{equation}    
    \mathcal{L}_{\text{axis}} = \left(|\cos\theta| + |\sin\theta\sin\phi| + |\sin\theta\cos\phi|\right)/3.
    \label{eq:loss_axis_reform}
\end{equation}
Taking partial derivatives of $\mathcal{L}_{\text{axis}}$ yields:
\begin{equation} 
\begin{aligned}    
    \nabla_{\phi}\mathcal{L}_{\text{axis}} &= \sin\theta\left(\pm \cos\phi \mp \sin\phi\right)/3,\\
    \nabla_{\theta}\mathcal{L}_{\text{axis}} &= \left(\mp \sin\theta \pm \cos\theta\left(\sin\phi + \cos\phi \right)\right)/3.   
\end{aligned}
\end{equation} 
Since $\mathcal{L}_{axis}$ yields local minima or maxima where $\nabla\mathcal{L}_{axis}(\theta, \phi)=0$,
solving the equation yields:
\begin{equation} 
\begin{aligned}
\nabla_{\phi}\mathcal{L}_{\text{axis}}=0 \Rightarrow \theta=0\text{ or }\phi=\frac{\pi}{4},\\
\nabla_{\theta}\mathcal{L}_{\text{axis}}|_{\phi=\frac{\pi}{4}}=0 \Rightarrow \theta=\arctan{\sqrt{2}}.
\end{aligned}
\end{equation} 
This analysis suggests that the global extrema are located at $\theta=0$ or 
$\left(\theta, \phi\right)=\left(\arctan\sqrt{2}, \frac{\pi}{4} \right)$. 
Substituting $\theta=0$ to $\mathcal{L}_{\text{axis}}$ yields $\mathcal{L}_{\text{axis}} \approx 0.333$, 
and $\left(\theta, \phi\right)=\left(\arctan\sqrt{2}, \frac{\pi}{4} \right)$ yields $\mathcal{L}_{\text{axis}} \approx 0.577$. 
Thus, we conclude that the $\mathcal{L}_{\text{axis}}$ reaches its minimum when $\theta=0$, 
indicating perfect alignment between the normal axis and one of the covariance axes when the loss is minimized.
Once $\theta$ reaches zero, the $\phi$ value becomes irrelevant, as the axis will align with the normal vector for all $\phi$.


\begin{figure}[!t]
    \centering
    \includegraphics[width=\textwidth]{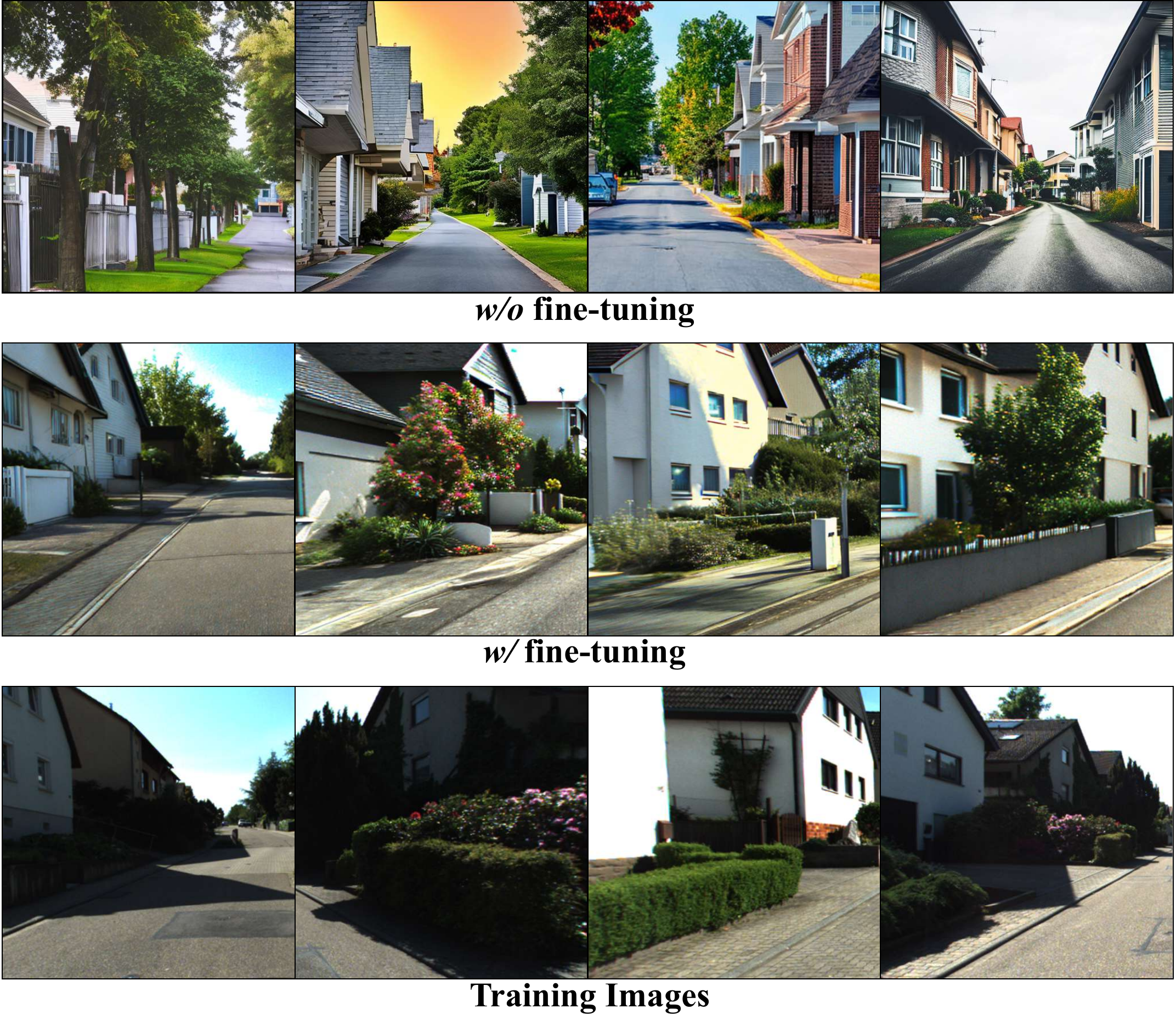}
    \vspace{-0.6cm}
    \caption{\textcolor{black}{Comparing samples generated with (\textit{top}) Stable Diffusion v2.1 \cite{Rombach_2022_CVPR}}, (\textit{middle}) our fine-tuned model, and (\textit{bottom}) training images of the scene. Fine-tuning model with LoRA \cite{hu2021lora} increases the scene-specific knowledge by large margin. For all sampling, we used the text \textit{"a photography of a suburban street"}.}
    \label{fig:sample}
\end{figure}

\section{Optimal Solution of Covariance Scale Loss}
In this section, we aim to substantiate that 
the covariance scale loss $\mathcal{L}_{\text{scale}}$ effectively minimizes the covariance scale along the covariance axis that is  most closely aligned with the normal axis.
By defining the covariance scale for each axis as $s_i$, 
and adhering to the angle notation in~\cref{eq:loss_axis_reform},
we can reformulate the covariance scale loss function~\cref{eq:loss_scale} as follows:
\begin{equation}    
    \mathcal{L}_{\text{score}} = \left(s_1|\cos\theta| + s_2|\sin\theta\sin\phi| + s_3|\sin\theta\cos\phi|\right)/3.
\end{equation}
In this formula, $\theta$ and $\phi$ are detached from the computational graph, 
indicating that they are only influenced by the covariance axes loss $\mathcal{L}_{\text{axis}}$.
Given that $\mathcal{L}_{\text{axis}}$ is designed to steer 
$\theta$ towards zero,
we can anticipate that $\mathcal{L}_{\text{scale}}$ will converge to the value of $s_1$ 
when the covariance axes loss functions as intended.
It is important to note that $s_1$ represents the scale along the covariance axis that aligns with the normal vector.
Therefore, minimizing $s_1$ is equivalent to flattening the covariance ellipse along the normal axis,
thereby aligning the covariance more closely with the surface and mitigating the cavity issue.
To prioritize the influence of the covariance axes loss over the covariance scale loss,
we assign 0.8 to $\lambda_{\text{axis}}$ in our experiments.

\section{Implementation Details}

\subsection{Training Details}
Our model is trained with 30,000 iterations, $\lambda_{\text{score}} = 10^{-11}$, $\lambda_\text{box} = 0.001$, $\lambda_{c}=1$, and $\lambda_{\text{axis}}=0.8$. Diffusion guidance is performed during the last 5,000 iterations in order to start guiding after $\mathcal{L}_{c}$ almost converges. 
To train VEGS with $\mathcal{L}_{\text{score}}$ loss, 
we use 512$\times$512 image, as the diffusion model is trained and best perform in 512$\times$512 image. 
Since the height of the KITTI-360 and KITTI dataset images are smaller than 512, we increase the image plane size to make its height 512, and random cropped by 512$\times$512 for diffusion score loss. Since our diffusion model assumes the total number of denoising step to be $T=1000$, we defined $\tau = 25$ to make it small enough to satisfy the assumption for \cref{eq:score}.

\begin{figure}[!t]
    \centering
    \includegraphics[width=0.5\textwidth]{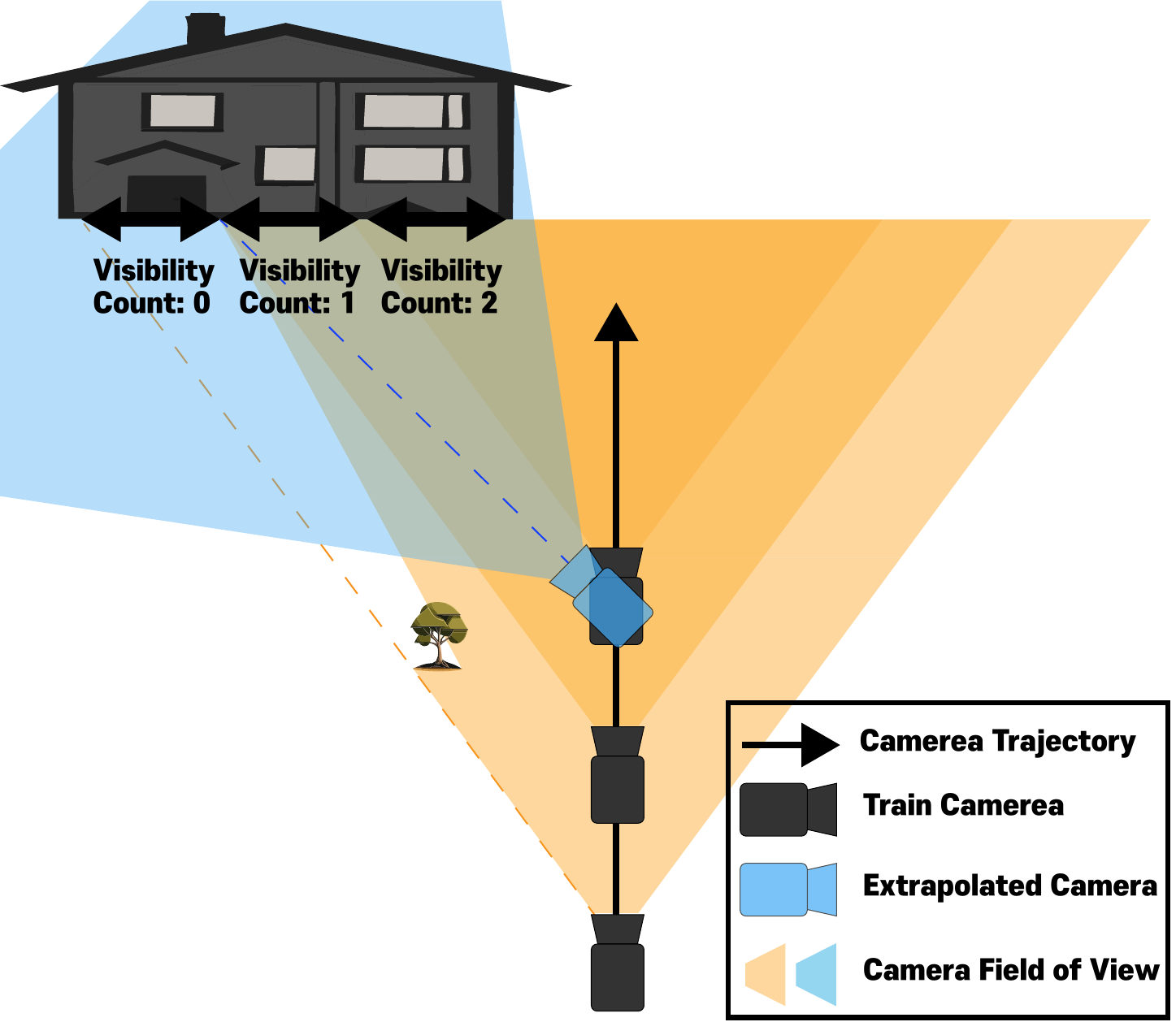}
    \vspace{-0.15cm}
    \caption{Illustration of a case where further half of EVS-LR observing occluded space.}
    \label{fig:visibility}
\end{figure}

\begin{table}
  \centering
   \resizebox{\linewidth}{!}{\begin{tabular}{x{3.5cm}|x{1.01cm}x{1.01cm}|x{1.2cm}x{1.2cm}x{1.4cm}x{1.4cm}}
    \toprule
     & FID $\downarrow$ & KID $\downarrow$ & PSNR $\uparrow$ & SSIM $\uparrow$ & LPIPS $\downarrow$ & PSNR* $\uparrow$ \\
    \midrule
    MARS & 209.61  & \sbest 0.167  &  \tbest 24.11 & \tbest 0.822  &  \tbest 0.119  &  \tbest 21.86  \\
    BlockNeRF++ & 394.71 & 0.342  & 23.56  & 0.789 & 0.172  & --  \\
    3DGS & \tbest 209.41 & \tbest 0.175 & 23.72 & 0.802 & 0.138 & --  \\
    3DGS+ & \sbest 182.87 & 0.207 & \best 24.82 & \best 0.847 & \sbest 0.115 & \best 23.06  \\
    VEGS  &  \best 167.38  & \best 0.090  & \sbest 24.77 & \sbest 0.845 & \best 0.113 & \sbest 23.01 \\
    \bottomrule
    \end{tabular}}
  \caption{Quant. comparison on KITTI \cite{Geiger2012CVPR}}
  \vspace{-0.7cm}
  \label{tab:kitti}

\end{table}

All experiments are conducted on RTX 3090 except for BlockNeRF++, which is trained on A6000 to handle VRAM of $\approx$ 48GB. We use omnidata\cite{eftekhar2021omnidata} for monocular normal estimation.
\subsection{Covariance Axis and Scale Initialization}
To ease the optimization process, 
we initialize the covariance axes and scales to align with our objective at initialization.
For the initial covariance axes, 
we project each point in LiDAR map to cameras to assign normal predicted from images to the point.
Since there are multiple normal vectors assigned to a point, 
we find a normal vector that is most likely to represent the normal of the point. To do so, we first construct intra-normal similarity matrix, followed by calculating the sum of similarity of a normal with respect to the other normals. Then, we select the normal that yields the highest similarity sum. Using the normal vector, we then establish the initial covariance axes by first defining one axis equal to the normal vector, another axis that is orthonormal to the first axis, and the last axis by applying Gram-Schmidt process to the first two axes, all of which consist a set of three orthonormal axes. 
As for initial covariance scales, 
we assign $1e\text{-}5, 1e\text{-}1, 1e\text{-}1$ to each axis, respectively.
We designate the smallest scale $1e\text{-}5$ to the axis that corresponds to the normal vector.

\subsection{Fine-tuning Diffusion Model} For large-scale diffusion model, we use Stable Diffusion v2.1\cite{Rombach_2022_CVPR}. As mentioned in our main paper, 
we fine-tuned the model using LoRA~\cite{hu2021lora} over 300 iterations, adopting a learning rate of $1\times10^{-4}$ and a cosine learning schedule.
For the text prompt $p$, we used \textit{"a photography of a suburban street"} for all experiments. 
Training images are randomly selected within the scene frame segment of interest. 
To prepare training dataset, we resized training images to have height of $512$ using bilinear interpolation. 
The training images are then copped at random positions by $512 \times 512$.

\subsection{Camera Resolution for Evaluation} As illustrated in
Fig.~\ref{fig:visibility}, the side region of EVS-LR plane 
inevitably has less observation due to the forward-facing nature of training cameras.  
Moreover, for the side regions that contain occluders, 
this lack of observation often leads to blank or noisy renderings.
This phenomenon arises regardless of recent methodologies. 
For this reason, to evaluate the regions that are properly reconstructed by multiple observations only, 
we crop the center of the frame for evaluation.



\section{Quantitative Results for KITTI}
We report the following experimental results on KITTI \cite{Geiger2012CVPR} in Tab. \ref{tab:kitti}, from which we can yield similar conclusion from KITTI-360.

\section{Additional Analysis}
\subsection{Covariance Visualization}

\begin{figure}[!h]
    \centering
    \includegraphics[width=\textwidth]{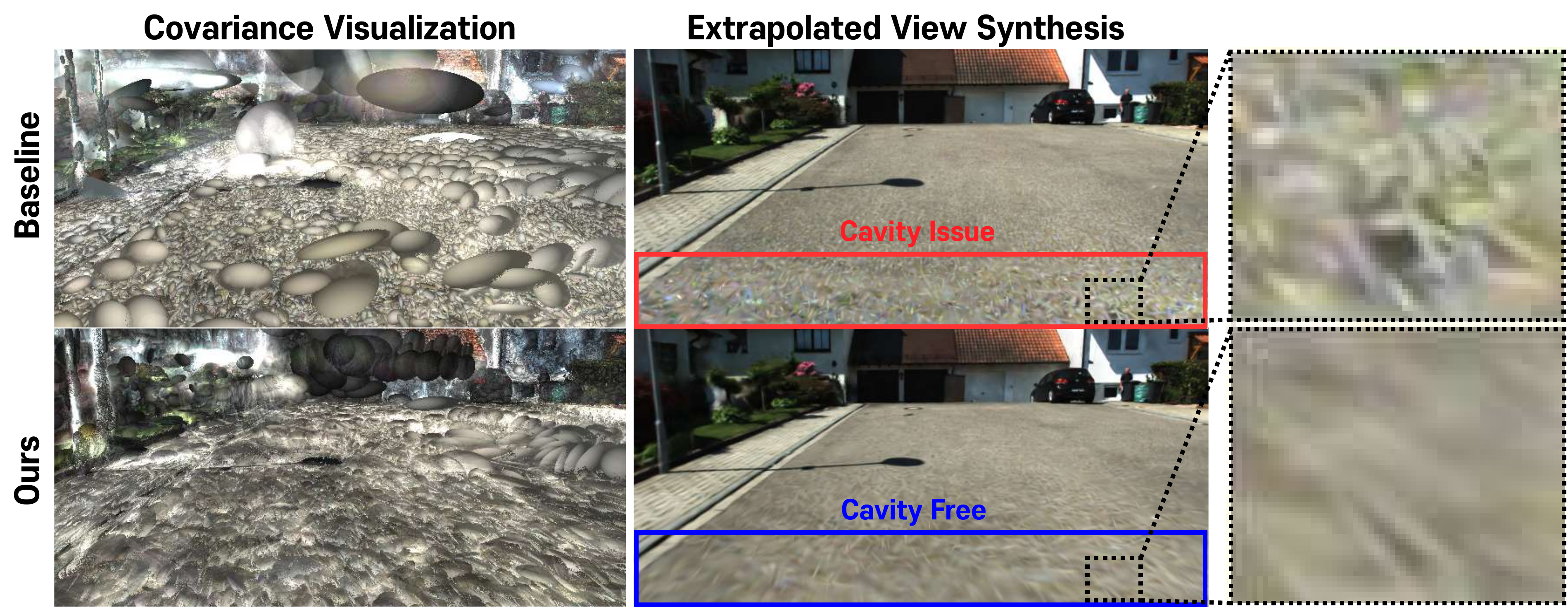}
    \vspace{-0.6cm}
    \caption{\textcolor{black}{Visualization of reconstructed covariance. The baseline method employs LiDAR points as the initial mean values for covariance estimation, omitting surface normal and diffusion priors. Our method ensures the alignment of one covariance axis with the surface normal, while simultaneously minimizing the covariance scale along this normal vector. Such alignment and scaling facilitates cavity-free extrapolated view synthesis.}}
    \label{fig:vis_cov}
\end{figure}

\begin{figure}[hbt!]
  \includegraphics[width=\linewidth]{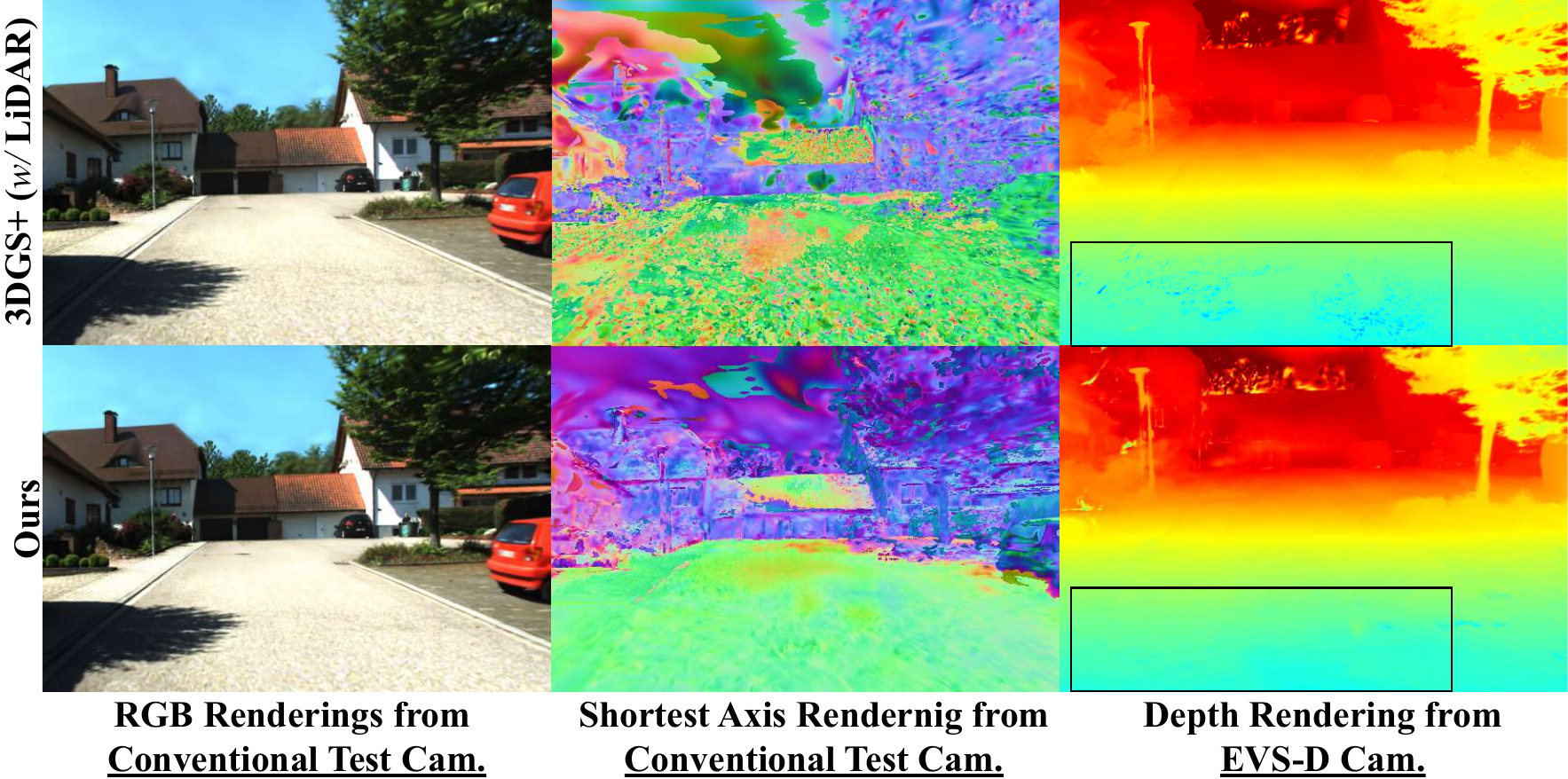}
  \vspace{-0.2cm}
  \caption{\textbf{(Left)} Renderings from a conventional test camera. \textbf{(Center)} Visualizing shortest axis of rendered covariance orientation map. \textbf{(Right)} Depth map rendered from EVS-D.}
  \label{fig:geo}
\end{figure}

In~\cref{fig:vis_cov}, we present a visualization of the covariances to illustrate the impact of the covariance axes loss and the covariance scale loss.
The figure demonstrates how our method 
not only aligns the covariance to the implicit surface normal but also effectively flattens it to encompass the surface comprehensively. 
This process enables the cavity-free extrapolated view synthesis by ensuring a seamless surface representation.
We also report the shortest covariance axis and depth renderings in Fig.~\ref{fig:geo} to visualize pseudo-normal and presence of cavity. 

\subsection{Effect of Stable Diffusion Fine-tuning} In order to verify that fine-tuning the model does represent the visual domain of the scene of interest, we generated samples from our model fine-tuned with LoRA, and compared with samples generated with the original pretrained model. We report the results in ~\cref{fig:sample}, which shows that samples generated with fine-tuned model look more visually close to the training images. 

\clearpage  
{\small
}


\end{document}